# Repurposing Annotation Guidelines to Instruct LLM Annotators: A Case Study

Kon Woo Kim[1,2][0000−0002−6970−5370], Rezarta Islamaj[3][0000−0001−5651−1860], Jin-Dong Kim[4][0000−0002−8877−3248], Florian Boudin[5][0000−0001−5849−2261], and Akiko Aizawa[1,2][0000−0001−6544−5076]

[1] The Graduate University for Advanced Studies, SOKENDAI
[2] National Institute of Informatics, Chiyoda, Tokyo, Japan
{ken1204,aizawa}@nii.ac.jp
[3] National Library of Medicine, Bethesda, MD, USA Rezarta.Islamaj@nih.gov
[4] Joint Support-Center for Data Science Research, Japan jdkim@dbcls.rois.ac.jp
[5] Japanese-French Laboratory of Informatics, CNRS, Nantes University, Japan
florian.boudin@univ-nantes.fr

**Abstract.** This case study explores the potential of repurposing existing annotation guidelines to instruct a large language model (LLM) annotator in text annotation tasks. Traditional annotation projects invest significant resources—both time and cost—in developing comprehensive annotation guidelines. These are primarily designed for human annotators who will undergo training sessions to check and correct their understanding of the guidelines. While the results of the training are internalized in the human annotators, LLMs require the training content to be materialized. Thus, we introduce a method called moderation-oriented guideline repurposing, which adapts annotation guidelines to provide clear and explicit instructions through a process called LLM moderation. Using the NCBI Disease Corpus and its detailed guidelines, our experimental results demonstrate that, despite several remaining challenges, repurposing the guidelines can effectively guide LLM annotators. Our findings highlight both the promising potential and the limitations of leveraging the proposed workflow in automated settings, offering a new direction for a scalable and cost-effective refinement of annotation guidelines and the following annotation process.

## 1 Introduction

Most annotation projects establish detailed annotation guidelines to effectively direct human annotators. In theory, well-crafted guidelines contain sufficient information to guide annotators' decisions even in the most nuanced cases, and the quality of the resulting annotations is closely tied to the clarity and precision of these guidelines. Consequently, many projects invest significant time and cost in developing labor-intensive yet comprehensive annotation guidelines, which serve as a highly valuable community resource. Incidentally, manual processes can often introduce unintended biases and human errors, producing inconsistencies in the finalized dataset, which could result in unexpected performance fluctuations of the downstream model [18]. To counteract this, researchers build annotation tools [11] and organize annotation tasks [13,12] to minimize unintended biases and human errors.

Meanwhile, large language models (LLMs) are increasingly recognized as viable alternatives to human labor across various domains and show the potential to assist human annotators and even automate at a significantly lower cost and in a much shorter time than manual annotation. Approaches to harness their capabilities range from zero-shot and few-shot methods to full fine-tuning [21,5,17,14].

In this case study, we investigate the potential of repurposing annotation guidelines to instruct LLMs in text annotation. As LLM capabilities increasingly approach those of human annotators, leveraging these extensive community resources becomes more promising. If LLM annotators can be effectively directed using existing guidelines, it would represent a significant benefit to the community. The study aims to evaluate this potential, assess the existing gaps, and offer insights into the remaining challenges in this area.

During the repurposing of annotation guidelines for LLM annotators, we focus on the moderation process typically used to train new human annotators. This training aims to bridge the gap between the intended interpretation of the guidelines and the annotators' understanding. To pinpoint this gap, trainee annotators first annotate sample texts, and any deviations from the gold annotations reveal areas that need improvement. While human annotators internalize this feedback, LLM annotators require training content to be explicitly materialized. We argue that the necessity for training underscores deficiencies in the original guidelines and that this training knowledge must be captured through revisions of annotation guidelines. In this work, we call this process moderation-oriented guideline repurposing. Our experiments on disease name annotations using the NCBI Disease Corpus reveal both the promising potential of employing annotation guidelines and the challenges that remain. This paper presents these findings in detail.

Our contributions are:

- We provide explicit instructions to an LLM annotator by repurposing existing domain-specific annotation guidelines.
- We introduce a workflow that mimics the human moderation process by automatically updating and adapting the guidelines for LLM annotations.
- We illustrate our approach to the NCBI disease corpus by applying the LLM annotation workflow to its annotation guidelines, which we repurposed for this study.
- We investigate the performance gap between LLM annotations and human annotations to show how the LLM annotator comprehends natural language differently in the annotation task.

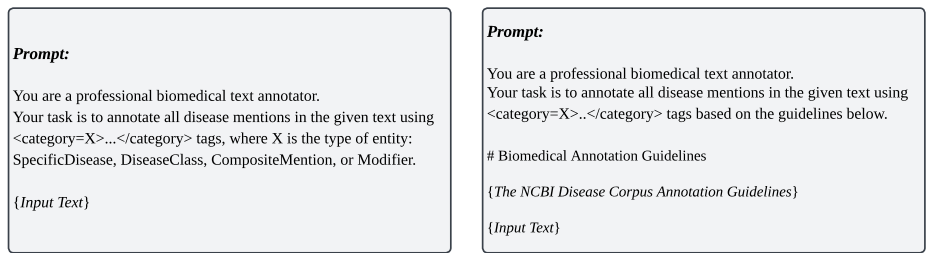

**Fig. 1.** Simple Prompting (Left) and Guidelines in the Prompt (Right)

## 2  Related Work

The use of crowd-sourcing workers has been a common practice to obtain natural language annotations until recently. However, studies have shown that the quality of annotated data from crowd workers is not guaranteed due to their limited expertise and knowledge[10]. As a result, researchers have explored alternative methods to ensure annotation quality [20,4,1]. To collect annotations more effectively, recent studies have focused on utilizing LLM annotators as an alternative to crowd workers or fine-tuned models. Recent studies suggest that LLM annotators have the potential to speed up the annotation process and even surpass the quality of crowd-sourced human workers for text annotation tasks. For example, ChatGPT showed text annotation performance comparable to both trained annotators and Mturk workers in certain datasets, including tweets, news articles, and FewRel [8,23,6], at lower costs. In addition, LLMs helped to improve the annotation process in certain domain-specific tasks [16] and fields, such as medical information extraction [9] and medicine [15], when properly aggregated with human annotators. A recent study attempted to utilize annotation guidelines to improve performance [19]. However, it still required some fine-tuning and did not fully utilize annotation guidelines because it only contained schema including label names, descriptions, and few-shot examples. In addition, it transformed annotation guidelines into a coding style, and therefore, it was difficult to adopt fine-grained annotation guidelines that cover various edge cases.

## 3  Methodology

To integrate comprehensive and detailed annotation guidelines for higher accuracy and consistency, we designed a workflow that the LLM annotator mimics the human moderation process using annotation guidelines and moderation process. Our methodology consists of several stages, progressing from a simple baseline to a full human-like annotation process.

### 3.1  Baseline Annotation

For the baseline, annotations were generated using a simple prompting method (as illustrated in Fig. 1). The prompt is divided into three parts:

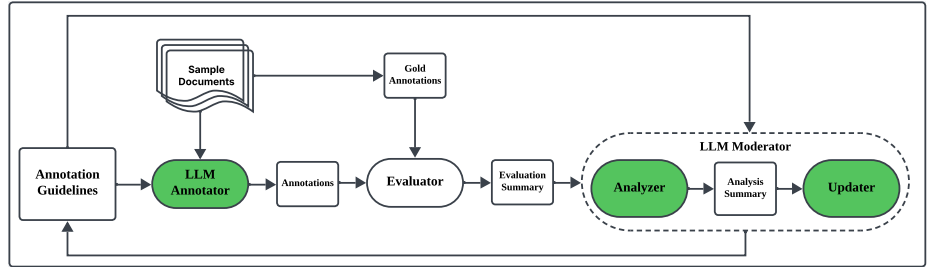

**Fig. 2.** The Workflow of Moderation-oriented Guideline Repurposing

1. A role definition as a text annotator to regulate the knowledge scope,
2. A brief description of the annotation task, and
3. Guidelines on output format and category, following the original dataset.

### 3.2  Guideline Integration

To validate the effectiveness of human annotation guidelines, we incorporated the annotation guidelines directly into the prompt. The remainder of the prompt remains the same as in the baseline to ensure a fair comparison as shown Fig. 1.

### 3.3  Moderation-oriented Guideline Repurposing

We created a workflow that mimics the human moderation process with the LLM annotator and LLM moderator to examine whether it can substitute the annotation task. In order to more closely mimic the human moderation process, we developed an iterative workflow with 3 phases (refer to Fig. 2). Following the general human annotation process, we began with the assumption that annotators annotate a small size of samples based on some initial gold annotations.

*Phase 1: Annotation*

– Provide a size of sample text documents from the dataset along with the original annotation guidelines formatted in a text file for improved readability.
– Instruct the LLM annotator to annotate the input text based on the provided guidelines.

*Phase 2: Evaluation and Summary*

– Compute the average F1-score in Strict Match against the gold annotations. If the average F1-score falls below threshold, the moderation process is initiated.
– Compare and extract discrepancies between LLM-generated annotations and gold annotations using the evaluator, including false positives (FPs), false negatives (FNs), and category mismatches.

*Phase 3: Moderation*

- Use the LLM moderator, which performs as both analyzer and updater, to analyze these discrepancies and generate a report that includes both the causes of the errors and potential solutions.
- Review the original guideline based on the generated report and update specific sections accordingly or provide detailed examples for edge cases.

### 3.4 Human-in-the-loop Moderation

Additionally, a human expert involved as an updater in Phase 3, conducted error analysis, and manually revised the guideline based on the LLM report to demonstrate the further potential of our proposed methodology.

## 4 Experiments

### 4.1 Dataset

To test our methodology, we selected the NCBI Disease Corpus [7]. It is comprised of 793 PubMed abstracts with 6,892 disease mentions and 790 unique disease concepts mapped to Medical Subject Headings (MeSH) or Online Mendelian Inheritance in Man (OMIM). Annotations are divided into four categories: Specific Disease(Mentions that can be linked to one specific definition that does not include further categorization), Modifier(A textual string may refer to a disease name, but it may modify a noun phrase), Disease Class(Mentions that could be described as a family of multiple specific diseases), and Composite Mention(A textual string may refer to two or more separate disease mentions). We selected this corpus for our experiment for two main reasons: 1. The NCBI Disease Corpus provides relatively shorter yet clear and structured annotation guidelines that can be used as a stepping stone to test our hypothesis. 2. It has been widely used in various model evaluations, such as BioBERT and SciBERT [14,3], which allows us to evaluate the significance and practical applicability of our proposed method.

### 4.2 Parameter Settings

For experiments, we selected GPT-4o as the LLM annotator and moderator, which is one of the most well-performing LLM models today. The size of sample text documents in Phase 1 is set to 5 to (i) loosen the burden of LLM moderator's revision complexity and (ii) account for real-world cases where only a small amount of gold annotation is available. The threshold used to break the loop in Phase 2 is set to 0.8, motivated by the lowest inter-annotator agreement observed in the NCBI Disease corpus.

### 4.3  Evaluation Criteria

We tested four evaluation criteria: Strict Match, Strict Match without Category Matching, Soft Match, and Soft Match without Category Matching. We also included Match without Category to measure the performance of disease name annotations without considering category confusion, reflecting requirements of downstream applications (e.g., PubTator [22]). We included Soft Matches as evaluation criteria not to show score improvements but to demonstrate the insufficiency of the span offset regulation when the LLM annotator is given annotation guidelines. A prior research also demonstrated that boundary free approach can improve inter-annotator agreement and reduce required annotation time[2]. Our criteria also provide insights into how well category handling is performed. Another reason is that strict boundary matching for annotation tasks in real-world cases is often challenging unless human judgments are moderated well, even if fine-grained annotation guidelines are provided.

## 5  Results

The results in Table 1 show that incorporating annotation guidelines significantly improves the accuracy of LLM annotations. Overall, the F1-score in the Strict Match evaluation improved from 0.36 to 0.58. Given the annotation guidelines, as shown in Table 2, the LLM annotator achieved 0.70 on Specific Diseases. Table 2 shows that the performance of the LLM annotator improved in Specific Diseases significantly as this category had clearer annotation guidelines requirements compared to the other categories. Incorporating the annotation guidelines outperforms the Baseline method in all criteria. This suggests that disease annotations can be better directed toward the gold annotations by instructing the LLM annotator based on the annotation guidelines used for generation of the gold annotations. The big gap between the Strict and Soft Match results indicates that precisely regulating the selection of the text spans to be annotated as disease names is still challenging. For all methods, matches without category tend to yield better F1-scores. This gap between with and without category matching indicates that category confusion, such as Disease Class vs. Specific Disease vs. Modifier, can be a major issue for the LLM annotator in a situation where category distinction is explicitly required.

The human-in-the-loop approach shows noticeable improvements, indicating that is still a room for more refinement for the LLM moderators. It can reduce category errors with human judgment in correctly defining detailed categories for the LLM annotator. To be more specific, Composite Mentions typically requires more complex spans with multi-words or nested concepts. For Disease Class, LLMs may do comparatively better if the prompts or training examples highlight unseen higher-level terms. The LLM annotator frequently gets confused with modifiers due to the inability to distinguish when certain terms can fit into multiple categories or failure to label them separately. The performance can be even stronger if it recognizes these disease mentions well, but boundary mismatching can penalize Strict Match heavily.

**Table 1.** Comparison of Annotation Performance under Different Matching Criteria. P: Precision, R: Recall, F1: F1-score. w/o Cat.: Without Category Matching.

| Method | Strict Match | | | Strict Match (w/o Category) | | | Soft Match | | | Soft Match (w/o Category) | | |
|---|---|---|---|---|---|---|---|---|---|---|---|---|
| | P | R | F1 | P | R | F1 | P | R | F1 | P | R | F1 |
| Baseline | 0.38 | 0.35 | 0.36 | 0.59 | 0.54 | 0.57 | 0.50 | 0.46 | 0.48 | 0.77 | 0.70 | 0.74 |
| GuidelinePrompt | 0.57 | 0.39 | 0.46 | 0.82 | 0.57 | 0.67 | 0.66 | 0.46 | 0.54 | 0.94 | 0.66 | 0.78 |
| Our Method | 0.56 | 0.45 | 0.50 | 0.82 | 0.65 | 0.73 | 0.62 | 0.49 | 0.55 | 0.90 | 0.72 | 0.80 |
| Human-in-the-loop | 0.67 | 0.51 | 0.58 | 0.82 | 0.62 | 0.71 | 0.72 | 0.55 | 0.62 | 0.90 | 0.69 | 0.78 |

**Table 2.** Comparison of Annotation Performance by Category. P: Precision, R: Recall, F1: F1-score. w/o Cat.: Without Category Matching.

| Category | Method | Strict Match | | | Strict Match (w/o Cat.) | | | Soft Match | | | Soft Match (w/o Cat.) | | |
|---|---|---|---|---|---|---|---|---|---|---|---|---|---|
| | | P | R | F1 | P | R | F1 | P | R | F1 | P | R | F1 |
| Composite Mention | Baseline | 0.08 | 0.35 | 0.14 | 0.40 | 0.35 | 0.37 | 0.14 | 0.60 | 0.23 | 0.66 | 0.70 | 0.68 |
| | Guideline | 0.36 | 0.50 | 0.42 | 0.57 | 0.65 | 0.61 | 0.50 | 0.70 | 0.58 | 0.86 | 0.90 | 0.88 |
| | Our Method | 0.24 | 0.25 | 0.24 | 0.62 | 0.40 | 0.49 | 0.43 | 0.45 | 0.44 | 0.90 | 0.65 | 0.76 |
| | Human-in-the-Loop | 0.06 | 0.10 | 0.07 | 0.47 | 0.10 | 0.17 | 0.22 | 0.40 | 0.29 | 0.78 | 0.50 | 0.61 |
| Disease Class | Baseline | 0.29 | 0.18 | 0.22 | 0.57 | 0.35 | 0.43 | 0.47 | 0.30 | 0.37 | 0.78 | 0.55 | 0.65 |
| | Guideline | 0.44 | 0.13 | 0.20 | 0.81 | 0.29 | 0.43 | 0.58 | 0.17 | 0.27 | 0.94 | 0.36 | 0.53 |
| | Our Method | 0.35 | 0.25 | 0.29 | 0.70 | 0.38 | 0.49 | 0.44 | 0.31 | 0.37 | 0.88 | 0.46 | 0.61 |
| | Human-in-the-Loop | 0.32 | 0.14 | 0.20 | 0.58 | 0.19 | 0.29 | 0.42 | 0.18 | 0.25 | 0.83 | 0.26 | 0.40 |
| Modifier | Baseline | 0.07 | 0.04 | 0.05 | 0.14 | 0.53 | 0.22 | 0.10 | 0.06 | 0.07 | 0.26 | 0.64 | 0.36 |
| | Guideline | 0.39 | 0.15 | 0.22 | 0.75 | 0.56 | 0.64 | 0.43 | 0.17 | 0.24 | 0.89 | 0.60 | 0.72 |
| | Our Method | 0.33 | 0.08 | 0.12 | 0.65 | 0.62 | 0.64 | 0.38 | 0.09 | 0.14 | 0.77 | 0.67 | 0.72 |
| | Human-in-the-Loop | 0.61 | 0.44 | 0.51 | 0.77 | 0.65 | 0.70 | 0.64 | 0.47 | 0.54 | 0.82 | 0.67 | 0.74 |
| Specific Disease | Baseline | 0.52 | 0.53 | 0.52 | 0.74 | 0.60 | 0.66 | 0.66 | 0.68 | 0.67 | 0.92 | 0.77 | 0.84 |
| | Guideline | 0.62 | 0.56 | 0.59 | 0.85 | 0.63 | 0.72 | 0.72 | 0.65 | 0.68 | 0.96 | 0.74 | 0.84 |
| | Our Method | 0.62 | 0.67 | 0.65 | 0.86 | 0.74 | 0.79 | 0.67 | 0.72 | 0.70 | 0.92 | 0.80 | 0.86 |
| | Human-in-the-Loop | 0.78 | 0.64 | 0.70 | 0.89 | 0.72 | 0.80 | 0.83 | 0.68 | 0.75 | 0.95 | 0.79 | 0.86 |

## 6 Discussion

We conducted an error analysis of the LLM annotator results in collaboration with humans and the LLM. We compared gold annotations with the LLM-generated annotations, identified FPs, FNs, and mis-categorized annotations, and the annotations categorized into two categories - that is in our case: disease and uncertain. Through the error analysis, we found two main reasonable causes resulting in unexpectedly lower F1-scores when incorporating annotation guidelines with the LLM annotator - scope mismatch and category mismatch.

### 6.1 Scope Mismatch

When NCBI Disease corpus annotators were asked to annotate disease names, first they had to define the scope of the disease. For example, they included specific disease names, "liver disease", "Wilson's disease", "colorectal cancer", etc., general disease references, "inherited disorder", "autosomal dominant disease", "neuromuscular disorder", etc., as well as phenotypic terms, such as "hyperthermia", "iron overload", "hepatomegaly", "clinodactyly", "broad great toes", etc. Annotation projects often adopt specific terminologies or ontologies to base their definition of the scope. In the case of NCBI Disease Corpus [7], it was MeSH and OMIM that defined the scope of disease names to be annotated.

**Table 3.** Category of Influencing Factors in the Dataset

| Category | Description | Examples |
| --- | --- | --- |
| Ambiguous Abbreviations and Acronyms (50.9%) | Short forms with multiple interpretations. Variability in capitalization or punctuation can contribute to ambiguity. | *DM* (diabetes mellitus or myotonic dystrophy), *CT*, *CD*, *IVF*, *A-T*, *DMS*, *PWS*, *TSD* |
| Generic or Vague Descriptors (17.7%) | Terms describing general concepts. | *tumor*, *tumours*, *tumors* |
| Chromosomal and Genomic Anomaly Terms (4.8%) | Genetic phenomena that are related to diseases. | *uniparental disomy*, *Maternal disomy 15*, *maternal UPD 15*, *UPD* |
| Descriptive or Phenotypic Features (10.0%) | Clinical descriptors relating to disease phenotypes. | *early death*, *premature death*, *facial asymmetry*, *broad great toes*, *limb anomalies* |
| Incomplete or Non-Specific Genetic/Pathological Descriptions (2.4%) | Phrases indicating abnormalities. | *genetic defect*, *gene defects*, *deficient activity of coproporphyrinogen III oxidase* |
| Miscellaneous or Low-Frequency Terms (14.2%) | Rare terms that may denote symptoms, components of a syndrome, or misparsed fragments. | *hypomania*, *mania*, *MJD*, *contractures*, *congenital joint contractures* |

According to the NCBI Corpus, some of the most common disease mentions are as follows: Tumor (43), DM (39), G6PD deficiency (33), APC (29). However, we noticed that the most common disease mentions were not likely to be annotated by the LLM annotator. According to the error analysis, influencing factors are categorized into 6 types as shown in Table 3. The LLM annotator skips the annotation of disease mentions specifying gene names. For example, in the NCBI Disease Corpus, "adenomatous polyposis coli" is annotated as a disease mention because this term is listed as the MeSH heading for the "Familial Adenomatous Polyposis Coli" disease. However, the LLM annotator would struggle with this term, which is mostly known to refer to the gene name causing the disease. In the articles annotated in the NCBI Disease Corpus, the term "adenomatous polyposis coli" or its abbreviation "APC" are annotated as "Disease Modifier" since they are often found to refer to the gene, i.e., APC gene. However, if human expert annotators were to re-annotate the NCBI Disease Corpus today, terms such as *adenomatous polyposis coli*, which can refer to both a gene and a disease, would likely be prioritized under the gene tag, since the current NLP systems are able to perform multiple entity recognition tasks concurrently. Another example that is often mis-annotated by the LLM annotator is the term "tumor(s)" or its alternate spelling "tumour(s)". "Tumor" refers to an anatomical mass and could be considered as a biological symptom, compared to "cancer", which refers to the diseased state. However, in the MeSH terminology, "tumor" and "cancer" are both listed as synonyms for the term "Neoplasms" categorized as a disease, and annotated as such in the NCBI disease corpus.

Myotonic Dystrophy (DM), however, is a case of ambiguity. The LLM annotator in this case expects the abbreviation of Myotonic Dystrophy to be MD. However, the developers of MeSH terminology aimed to reduce ambiguity by not assigning the same text string to denote two different concepts. As such, since there already existed a concept named Muscular Dystrophy abbreviated "MD", Myotonic Dystrophy was referred to by its Greek name - dystrophia myotonica and abbreviated DM.

### 6.2 Category Mismatch

Category mismatches occur when the LLM annotators have difficulties in understanding the nuances of the annotation guidelines. For the case of NCBI Disease Corpus, the "Composite Mention" category defines a mention that refers to more than one disease, for example: "Saethre-Chotzen, Crouzon, and Pfeiffer syndromes". In this case, the LLM annotator only annotated "Pfeiffer syndromes" as a Specific Disease, creating a category mismatch. If these mentions were written in full rather than abbreviated with ellipsis, then all three would have been annotated as Specific Disease. Furthermore, the "Modifier" category defines any disease mention that is used as an adjective or modifies a noun, for example "APC" gene, "DM" mutation, "breast cancer" families, "aniridia" patients. The LLM annotator was more likely to skip mentions in the modifier context, especially if they referred to genes. Lastly, "Disease Class" denotes a mention that refers to a group of diseases, rather than a specific one, for example: "adenomas", "autosomal dominant disease", "cancer", "leukemia", etc. The LLM annotators had difficulties distinguishing between Specific Diseases and General Diseases.

## 7 Conclusion and Future Work

The performance of the LLM annotator gradually increased from 0.36 to 0.58 in the Strict Match F1-score after repurposing the annotation guidelines. Our moderation-oriented guideline repurposing involves two steps - 1. incorporating the annotations into the prompt, which allows easier modifications, and 2. moderation-oriented guideline revision, which better instructs the LLM annotator towards the gold annotations. This improvement required less time and cost compared to fine-tuning methods, which solely focuses on instructing the LLM annotator for an annotation task. In addition, LLM annotators have the potential to explain and adapt the latest scope of category definitions capturing the evolving landscape of language use. This can open new avenues for the use of LLMs in repurposing old annotation guidelines. As a future work, we expect to (i) allow the LLM to specifically use the original ontology/terminology as a reference for annotation while following the annotation guidelines; (ii) improve the moderation process so that we can improve the accuracy of categorization and reduce human involvement as we gradually move towards fully automating the workflow for higher-quality annotations; and (iii) expand this study to other datasets and domains to validate its practicality.

**Acknowledgments.** This work was supported by Cross-ministerial Strategic Innovation Promotion Program (SIP) on "Integrated Health Care System" Grant Number JPJ012425, and JSPS KAKENHI Grant Number 24K03231.